\documentclass[runningheads]{llncs}
\usepackage[T1]{fontenc}
\usepackage{graphicx}
\usepackage{censor}
\usepackage{xurl}
\usepackage{url}
\usepackage{hyperref}
\begin{document}
\title{Known-Voice Cloning and the Limits of Personality Transfer for Character Design\thanks{This preprint has been accepted for publication in \textit{Connected Futures – Nature, Robots, and Society} (Robophilosophy 2026), in the \textit{Lecture Notes in Computer Science} (LNCS) series, volume 17299 (expected publication: December 2026).}}
\titlerunning{Known-Voice Cloning}

\author{anonymous authors}
\institute{anonymous institute}
\author{
Johanna Magdalena Kuch\inst{1}\orcidID{0000-0001-5322-9552} \and\\
Maud van Lier\inst{2}\orcidID{0000-0001-7951-7026} \and\\
Daksitha Senel Withanage Don\inst{1}\orcidID{0009-0003-8831-9795} \and\\
Patrick Gebhard\inst{3}\orcidID{0000-0002-5566-4520} \and\\
Tanja Schneeberger\inst{3}\orcidID{0000-0002-7247-6978} \and\\
Elisabeth André\inst{1}\orcidID{0000-0002-2367-162X}
}

\authorrunning{Kuch et al.}

\institute{
Augsburg University, Human-Centered Artificial Intelligence, Augsburg, Germany \\
\email{\{johanna.kuch,daksitha.withanage.don,elisabeth.andre\}@uni-a.de}
\and
Institute of Intelligent Systems and Robotics, CNRS, Sorbonne Universit\'e, Paris, France
\and
German Research Center for Artificial Intelligence (DFKI), 
Affective Computing Group, Germany
}

\maketitle              

\begin{abstract}
Recent advances in zero-shot voice cloning systems make it possible to reproduce a recognizable voice from a short recording and attach it to arbitrary virtual characters. Prior work suggests that familiar vocal identity can carry over associated personality impressions to new targets such as robots or virtual characters. However, these works have mainly examined familiarity and personalization effects rather than directly testing whether a cloned familiar voice systematically transfers personality impressions to an embodied virtual character. Such transference raises the concern that cloned voices could be used to manipulate people into engaging with virtual characters as capable social actors. Across two studies, we investigated whether personality impressions transfer when a virtual character speaks with the cloned voice of a familiar person. Study 1 examined this question in a real-world university setting using the cloned voice of a lecturer familiar to the participating students. Study 2 examined the same question online using voices of social media content creators and members of their communities. Drawing on the concepts of identity activation, trait transfer, and person reconstruction, we show that evidence of personality transfer was limited in both studies. Observed effects were selective and did not generalize across traits or study contexts. In the online study in particular, recognizable voice identity cues activated expectations about the speaker but did not reconstruct the broader person impression. Qualitative responses further indicated that judgments were frequently shaped by perceived artificiality and by whether voice and appearance fit together. Our findings moderate common concerns that cloned voices straightforwardly manipulate people into perceiving artificial characters as capable social actors.



\end{abstract}
\section{Introduction}

From short speech samples, listeners infer personality traits and social qualities such as trustworthiness, dominance, and competence \cite{McAleer.2014}. When a voice is familiar, perception extends beyond acoustic inference: Familiar voices are stored in memory with sufficient specificity for robust identity recognition and activate stored person-specific knowledge, including trait attributions and relational associations \cite{Kanber.2022}.

Research has now shown that this stored knowledge can be activated on the basis of particular cues alone, leading us to project person-based attributions from familiar individuals onto new targets, such as when we feel an instant liking for a new person who resembles a dear friend in some way \cite{Kraus.2010,TalOr.2021}. Since hearing a familiar voice can trigger person-specific representations or associated attributions even without explicit identification \cite{Campeanu.2015,Lavan.2025}, it is likely that familiar voices can form such cues as well. If they can, then it is possible that the attachment of familiar voices to new targets can bias our judgments of these targets. 

From a social robotics and AI perspective, this is an interesting possibility, especially now that recent work suggests that we evaluate synthetic voices along similar social dimensions as we do natural human voices \cite{Shiramizu.2022}. If we already infer personality traits and social qualities from synthetic voices, what would happen when we provide robots or virtual characters not just with human voices, but with voices of humans that we are already familiar with? Would it activate our stored person-knowledge, biasing our evaluation of the capacities of that system to function as a social actor, thereby increasing our willingness to trust and engage with that system?

Prior work has demonstrated that voice design can shape the appeal, trustworthiness, and perceived believability of virtual agents and characters \cite{Chiou.2020,Higgins.2022}. And, even when just a synthetic voice is used, similarity-based voice design has been shown to increase robot likability (both with and without conscious awareness of the similarity) \cite{Kuch.2025}. In addition, related work suggests that character-inspired voices can shape user engagement and evaluations of conversational AI systems \cite{Figueroa.2026}. Based on these findings, the hypothesis can be drawn that the use of a familiar voice is likely to further increase our willingness to engage with an artificial system. What this seems to depend on, however, and what the aforementioned findings do not yet establish, is whether a cloned familiar voice can systematically transfer personality impressions to an embodied artificial character and whether the coupling of a known voice with a visible body amplifies such transfer.  

This is an important question to answer, especially given recent advances in zero-shot voice cloning. Such systems generate realistic speech from short reference recordings \cite{Casanova.2024,Qin.2023}, making it possible for recognizable voices to be reproduced and attached to arbitrary characters across a variety of contexts. Besides the ethical concerns raised by the fact that there are not yet adequate policies in place to protect us from having our voices cloned and used without consent,\footnote{Voice cloning detaches vocal identity from the embodied person and makes its reuse scalable across platforms. Public debates about deepfakes and voice cloning highlight perceived threats to identity and trust. Beyond authentication risks \cite{Wang.2025}, qualitative research documents fears of identity displacement, fragmentation of selfhood, and loss of control over vocal representation \cite{Du.2025,Lee.2023,Leuenberger.2025}.\label{footnote4}} there is also the worry that the implementation of familiar voices in Human-Robot and AI interaction can constitute a \textit{dark pattern} \cite{Shamsudhin.2021}.\footnote{We adopt here the definition of a dark pattern proposed by Shamsudhin \& Jotterand \cite{Shamsudhin.2021}, who define a dark pattern as a persuasive design of which ``(1) the intent to persuade must be necessary, (2) the resulting user action is not in the user's best interest and is in the interest of the persuader, and (3) the methods employed are deceptive and covert'' (p.~96).} After all, if familiar vocal identity indeed carries over associated personality impressions in a reliable way, then designers could strategically activate character traits through voice selection alone, thereby manipulating users to engage with their artificial systems in ways that are not necessarily in their best interest. Given these pressing ethical concerns, it is thus important that we find out whether voice-based familiarity provides a plausible pathway for attribution transfer. 

In this paper, two studies are presented that provide initial insights into the possible effects that the use of familiar voices can have on our evaluation of an artificial system as a capable social actor. We thereby make three contributions to existing research: (1) We provide empirical evidence across two contexts that familiar voice cloning yields only small and trait-selective alignment effects rather than robust person-consistent evaluations. (2) We show, through qualitative analysis, that judgments are often dominated by perceived artificiality and voice--appearance mismatch. (3) We derive practical guidance for responsible character design by identifying boundary conditions under which familiar voice cloning is unlikely to support social evaluation and may backfire.

In the following, we first introduce our main concepts (section \ref{concepts}) and present the two studies and their results (sections \ref{method}--\ref{results}). Based on these results, we argue that familiar voices can influence evaluations of artificial systems as social actors, but that this influence is selective, context-dependent, and insufficient to support straightforward assumptions of robust personality transfer or person reconstruction. We end by discussing limitations of the present studies and outline directions for future work (sections \ref{discussion}--\ref{limitations}).

\section{Conceptual Framework}
\label{concepts}

To clarify the possible effects of familiar voices in robots and virtual characters, it is useful to distinguish between three conceptually different outcomes.

\textbf{Identity activation} occurs when a familiar voice triggers recognition or retrieval of a stored representation of the original speaker. Hearing the voice can activate associated person knowledge, expectations, and affective responses linked to that individual. Such activation may influence how the artificial character is perceived or evaluated. However, the influence remains indirect and does not necessarily result in consistent attribution of the original speaker's personality traits to the new target \cite{Lavan.2023}.

\textbf{Trait transfer} (as used in this paper) refers to a pattern in which evaluations of an artificial character converge toward trait and social judgments associated with a familiar speaker when the character uses a cloned version of that speaker’s voice. Importantly, this is not meant as \emph{projection} in the classic sense of attributing one’s own characteristics to others. Rather, it describes an \emph{attributional carryover} in which activated person knowledge about the familiar speaker systematically biases judgments of the new target.

\textbf{Person reconstruction} represents the strongest possible form of transfer. In this case, the familiar voice cue would reinstate a broader person impression, such that listeners evaluate the artificial character in ways that closely resemble their internal model of the original individual across multiple traits and social judgments \cite{Kraus.2010,Rosi.2025}.

These outcomes differ in strength and theoretical implication. \textit{Identity activation} concerns the retrieval of person-related expectations, \textit{trait transfer} concerns selective attribution of characteristics, and \textit{person reconstruction} would imply that an isolated perceptual cue can sustain a coherent person representation. The studies presented in this paper examine whether familiar voice cloning produces evidence for \textit{trait transfer} and whether it approaches the stronger case of \textit{person reconstruction} in evaluations of embodied artificial characters.

\section{Method}\label{method}

Table~\ref{tab:study_overview} summarizes the main design characteristics of the two studies.
We use the empirical studies as a case to examine how voice-based identity cues operate in embodied characters.

\begin{table}[b]
\centering
\caption{Overview of the two studies.}
\label{tab:study_overview}
\begin{tabular}{lcc}
\hline
 & \textbf{Study 1} & \textbf{Study 2} \\
\hline
Context & University lecture & Online communities \\
Participants ($n$) & 17 & 65 \\
Known-voice owner & Lecturer & Content creator \\
Design & Within-subject & Within-subject \\
Stimulus length & $\sim$14 s & $\sim$43 s \\
Speech content & Oldenburg sentences \cite{Kuehnel.1999} & Trivia statements \cite{Nadarevic.2022} \\
Personality measure & BFI-K (21 items) \cite{Rammstedt.2005} & BFI-10 \cite{Rammstedt.2014} \\
Likability \& Competence & \multicolumn{2}{c}{Items from \cite{Carli.1995,Hale.2016}; German adaptation \cite{Wessler.2022}} \\
Trust & -- & \cite{Roth.2018} \\
\hline
\end{tabular}
\end{table}

Participants first rated a familiar real person from memory to capture their baseline internal person model. They then evaluated the same virtual character in two voice conditions: a \textit{known-voice clone} and a \textit{cloned unknown voice}. Visual appearance and spoken content were held constant, making voice the only manipulated factor. The unknown voice was unfamiliar to participants and served as control voice for the same character and speech content.

If the \textit{known-voice clone} activates stored person knowledge and supports attribution transfer, character ratings should be closer to real-person ratings in the \textit{known-voice clone} condition than in the \textit{unknown voice} condition. We operationalized this as smaller absolute distances between known-person and character ratings across personality traits and social evaluation measures, indicating greater alignment between the remembered person model and the evaluated character.

The order of the two character conditions was randomized in Study~1 and Study~2; the memory-based familiar-person rating always preceded the character evaluations.

Study~1 was an exploratory study using a university lecturer as the familiar person. Participants were recruited in a university teaching context and completed the study during a lecture or tutorial using their own smartphones and headphones. The familiar person was the lecturer whose course they attended. Of 18 completed questionnaires, one participant was excluded for not knowing the lecturer, leaving 17 valid responses. Participants had a mean age of 22.19 years (\textit{SD} = 2.99; range: 20--32).

Study~2 tested the same design online using the voices of two social media content creators. Participants were recruited through creator-related online communities, primarily Reddit, supplemented by mailing lists and social media posts. Of 208 participants who started the survey, 65 valid responses remained after applying preregistered exclusion criteria, including incomplete responses, failed attention checks, and insufficient creator recognition. Participants had a mean age of 27.9 years (\textit{SD} = 6.38; range: 19--52).

At the beginning of Study~2, participants completed a familiarity check for the two selected creators and indicated which one they knew better. Actual familiarity was verified by requiring correct identification of both the creator's real name and artist name, as well as content consumption within the previous three months. Participants who failed these preregistered criteria were excluded. After rating the selected creator from memory, participants evaluated the virtual character in the known-voice and unknown-voice conditions.

Study~2 was preregistered.\footnote{\label{fn:prereg}\url{https://aspredicted.org/zy8j9m.pdf}} All participants took part voluntarily and provided informed consent before participation. Voices were cloned with permission from the respective voice owners or from materials explicitly authorized for the study.

\subsection{Measures}
\paragraph{Personality.}
Study~1 assessed Big Five other-ratings using the German Big Five Inventory adapted for rating others \cite{Rammstedt.2005,Wiechers.2025} (21 items; five-point Likert scales; negatively keyed items reverse-coded). Study~2 used brief Big Five scales adapted for other-ratings \cite{Rammstedt.2014}.

\paragraph{Social evaluation.}
Likability was measured with a multi-item scale \cite{Hale.2016,Wessler.2022}, and competence with established person-perception items \cite{Carli.1995}. Study~2 additionally measured trust \cite{Roth.2018}. Likability and competence used seven-point Likert scales in Study~1; Study~2 followed the instruments' original response formats.

\subsection{Stimuli}

\begin{figure}[t]
\centering
\includegraphics[height=3.5cm]{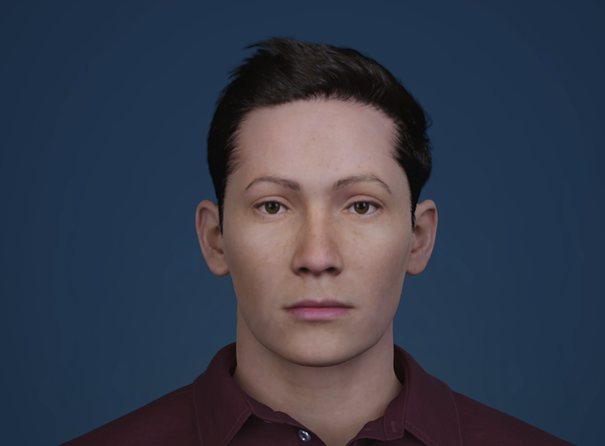}
\includegraphics[height=3.5cm]{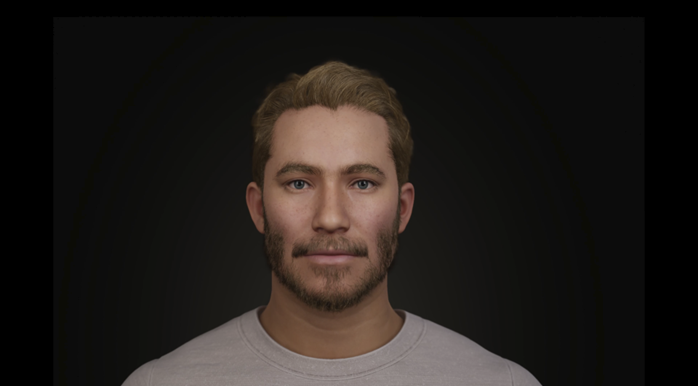}
\caption{Virtual characters used in the studies. Left: Character used in Study~1 (exploratory real-world study). Right: Character used in Study~2 (online social media study). In both studies, visual appearance and spoken content were held constant across voice conditions.}
\label{fig:characters}
\end{figure}

Two different virtual characters were used across the studies (Figure~\ref{fig:characters}). Within each study, visual appearance and spoken content were held constant across voice conditions. The order of voice conditions was randomized. 

Voice clones were generated from short reference recordings using zero-shot voice cloning systems \cite{Casanova.2024,Qin.2023}: OpenVoice in Study~1 \cite{Qin.2023} and XTTSv2 in Study~2 \cite{Casanova.2024}.

The virtual character and lip synchronization were created with MeLaX \cite{WithanageDon.2025}. Stimulus duration differed between studies (Table~\ref{tab:study_overview}): Study~1 used shorter, phonetically balanced sentences \cite{Kuehnel.1999}, whereas Study~2 used longer statements \cite{Nadarevic.2022} to support more reliable personality judgments following participant feedback from Study~1.

\subsection{Analysis}
Study~1 was exploratory and used paired t-tests or Wilcoxon signed-rank tests depending on the distribution of the difference scores. Study~2 followed the preregistered directional analysis plan, using one-tailed paired t-tests or one-tailed Wilcoxon signed-rank tests to test whether distances were smaller in the known-voice than in the unknown-voice condition. Distances were calculated as absolute differences between person and character ratings on each scale.

Open-ended responses from Study~2 were analyzed using qualitative content analysis \cite{Mayring.2015}, focusing on voice recognition, voice--appearance fit, perceived artificiality, and emotional tone.

Given its exploratory nature, Study~1 used no correction for multiple comparisons. Study~2 tested the directional hypothesis separately for each outcome according to the preregistered analysis plan,\textsuperscript{\ref{fn:prereg}} also without correction for multiple comparisons.

\section{Results}
\label{results}

\begin{figure}[!t]
\centering
\includegraphics[width=\linewidth]{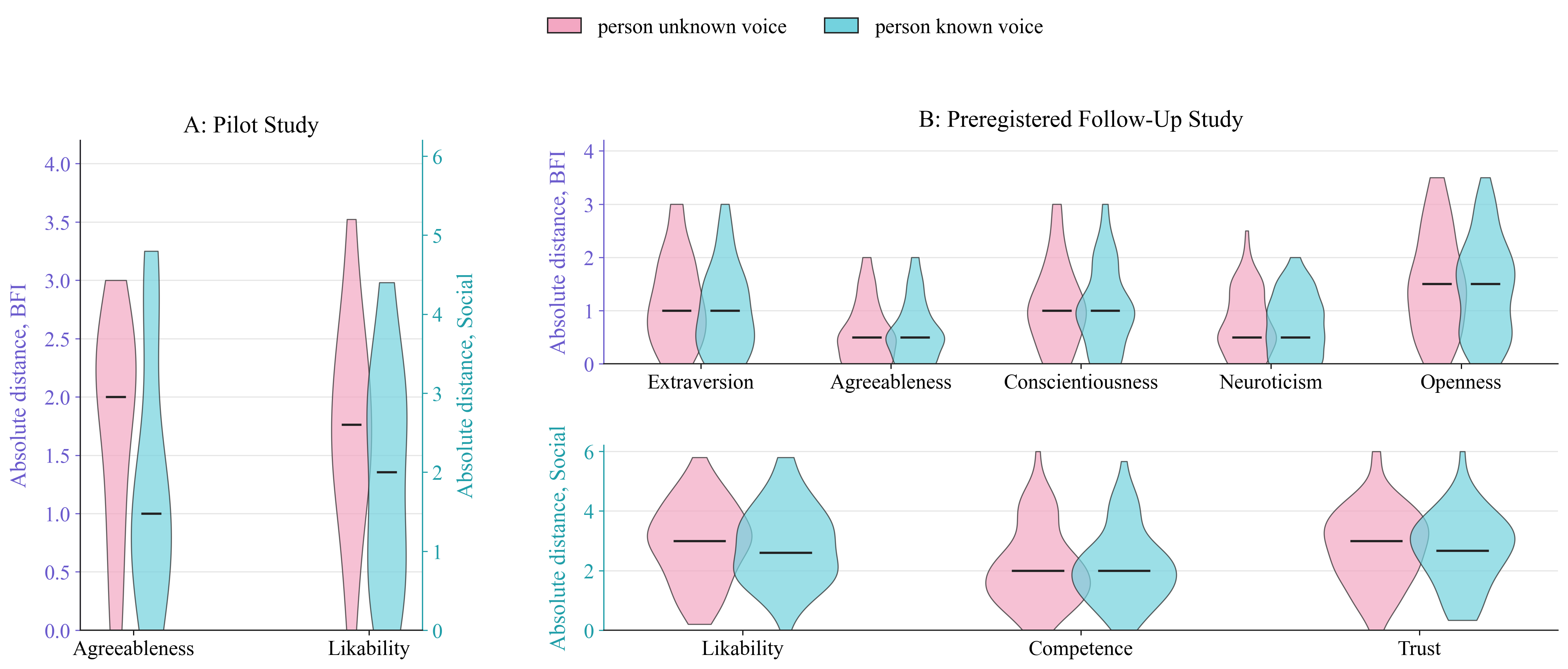}
\caption{Absolute distances between known-person ratings and character ratings for the unknown-voice and known-voice conditions. Panel A shows Study~1 ($n=17$); Panel B shows Study~2 ($n=65$). Violins show distance distributions, and horizontal lines indicate medians. In Panel A, BFI and social evaluation distances are shown on separate y-axes because the underlying response scales differed.}
\label{fig:combined_AB}
\end{figure}

\subsection{Study 1 (Lecturer, $n=17$)}
Wilcoxon signed-rank tests showed significantly smaller distances in the known-voice condition for likability ($W=11.5$, $p=.010$, rank-biserial $r=-.75$) and agreeableness ($W=16.0$, $p=.038$, rank-biserial $r=-.59$). Likability distances were lower in the known-voice condition ($M=1.96$, $SD=1.35$, $Mdn=2.00$) than in the unknown-voice condition ($M=2.56$, $SD=1.32$, $Mdn=2.60$); the same pattern was found for agreeableness ($M=1.28$, $SD=1.05$, $Mdn=1.00$ vs. $M=1.74$, $SD=0.92$, $Mdn=2.00$). Competence and the remaining Big Five dimensions were not significant (Figure~\ref{fig:combined_AB}).

\subsection{Study 2 (Content creators, $n=65$, preregistered)}
Only extraversion showed a significant effect in the predicted direction ($W=588.0$, $p=.041$), with descriptively lower distances in the known-voice condition ($M=1.02$, $SD=0.81$, $Mdn=1.00$) than in the unknown-voice condition ($M=1.19$, $SD=0.84$, $Mdn=1.00$). All other outcomes, including likability, competence, and trust, were not significant (Figure~\ref{fig:combined_AB}). Thus, the Study~1 effects for likability and agreeableness did not reappear in Study~2; instead, only extraversion showed the predicted effect.

\subsection{Qualitative Findings}\label{qualitativefindings}
\begin{figure}[h]
\centering
\includegraphics[width=0.75\linewidth]{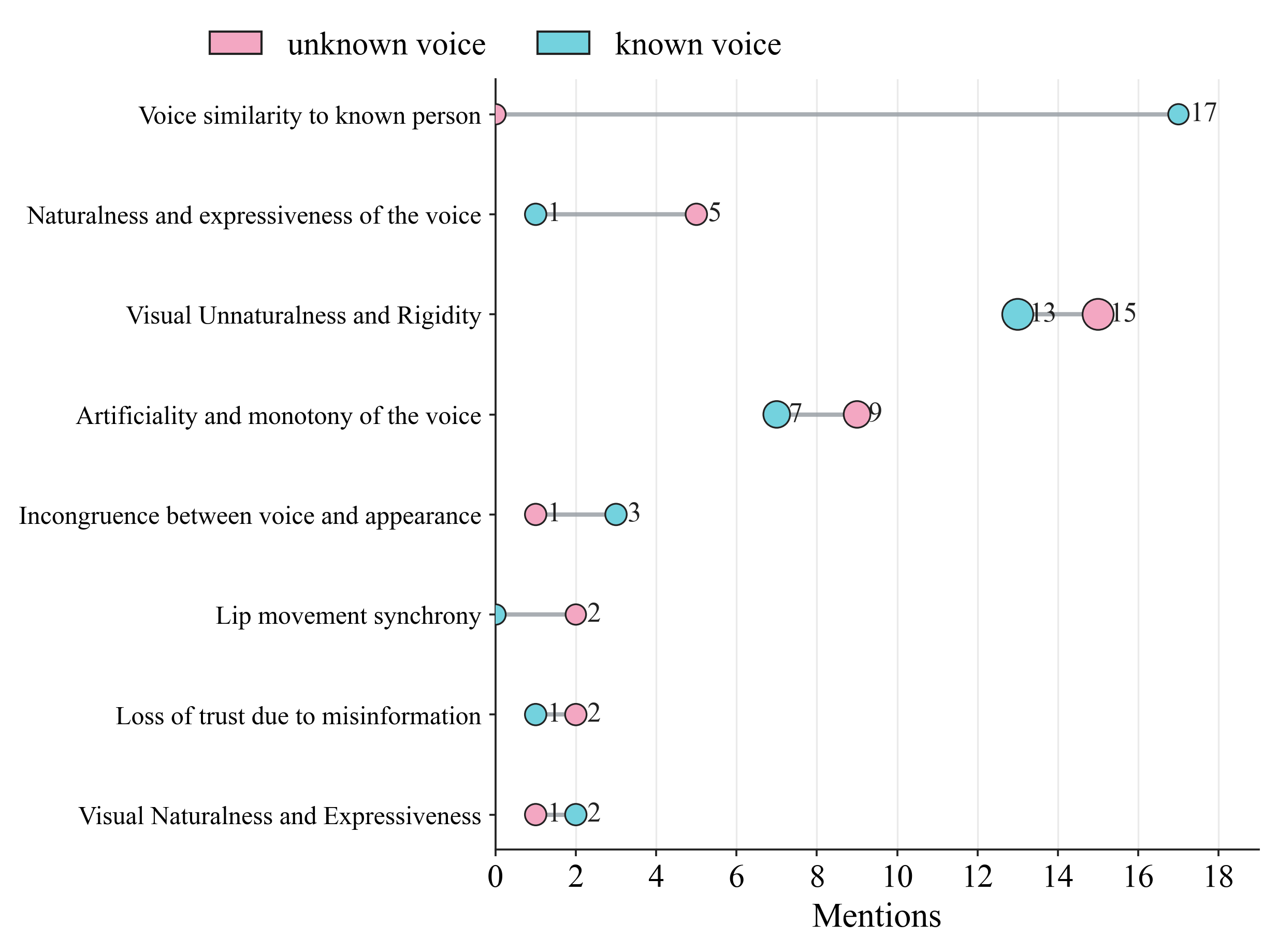}
\caption{Frequency of coded themes in open-ended responses by voice condition. Numbers indicate mentions.}
\label{fig:qualitative}
\end{figure}

Open-ended responses from Study~2 were analyzed using qualitative content analysis \cite{Mayring.2015}. Figure~\ref{fig:qualitative} shows the frequency of coded themes by voice condition.

Recognition of the cloned voice was salient in the known-voice condition (17 mentions). Participants often acknowledged similarity to the familiar person while emphasizing deviations from the original voice, indicating imperfect perceived fidelity.

Across both conditions, the dominant theme was visual rigidity and artificial facial movement (unknown: 15; known: 13 mentions). Participants described limited expression, stiffness, and, in some cases, an uncanny or unsettling impression, independent of voice familiarity.

Comments describing the voice as comparatively natural or pleasant were more frequent in the unknown-voice condition (5 vs.\ 1 mention), whereas voice--appearance mismatch was reported more often in the known-voice condition (3 vs.\ 1 mention).

Several participants questioned whether personality judgments were possible under the given conditions, suggesting perceived limits of trait attribution. In isolated cases, familiarity slightly attenuated negative impressions without reversing them. Distrust-related comments primarily referred to perceived misinformation rather than to the use of a familiar voice.

\section{Discussion and Ethical Implications}\label{discussion}

Across both studies, familiar voice cloning produced only limited alignment between character ratings and memory-based ratings of the original speaker. Effects were selective, context-sensitive, and did not generalize across traits or study contexts. Interpreted through the conceptual distinction introduced earlier, the findings are therefore most consistent with \textit{identity activation}, provide only weak evidence for \textit{trait transfer}, and provide little evidence for \textit{person reconstruction}.

The following sections discuss why familiar voices appear to function less as direct carriers of personality than as partial identity cues whose effects depend on the broader multimodal context.

\subsection{A weak transfer channel under cue competition}

Effects varied clearly by trait. Only extraversion\footnote{In personality psychology, extraversion refers to a tendency toward sociability, talkativeness, energetic behavior, and outward emotional expressiveness. In the Big Five framework, it is typically contrasted with introversion \cite{John.1999}.} showed the predicted uncorrected directional effect in Study~2, whereas agreeableness and likability effects from Study~1 did not replicate. This difference may relate to the nature of familiarity with the speaker. In Study~1, participants knew the lecturer from direct interaction, likely resulting in richer and more stable person representations that extend beyond vocal cues. In contrast, familiarity with social media creators is typically mediated and may rely more strongly on cues such as voice and visual appearance. Under such conditions, vocal cues may play a comparatively larger role in shaping trait impressions, which may partly explain why extraversion showed the clearest difference in the online context. Extraversion is closely tied to acoustic parameters such as vocal effort, intensity, and dynamic variability, making it comparatively voice-diagnostic \cite{Mohammadi.2012,Scherer.1978}. Traits such as agreeableness or likability may rely more strongly on broader behavioral cues and may therefore be less recoverable from voice alone.

At the same time, overall effect sizes were small. Qualitative responses indicate that perceived vocal artificiality and visual rigidity often shaped impression formation. Familiar listeners are known to hold relatively robust mental representations of familiar voices and can detect subtle deviations in prosody, timing, and accent \cite{Barrington.2025,Kanber.2022,Rosi.2025}. When a cloned voice diverges from this stored prototype, attention may shift toward evaluating the authenticity of the voice rather than forming person-based impressions. This may also affect social evaluations such as perceived warmth or credibility.

Prior work suggests that technical imperfections draw attention to a voice's artificiality \cite{Rosi.2025}. While it has been shown that familiar voices in emotionally charged contexts can retain persuasive effects despite such deficits \cite{Sharevski.2025}, these findings were based on cases where listeners were exposed to audio cues only, so without a visual carrier that embodied the voice. Our studies suggest that coupling a familiar voice with an embodied character produces mismatch awareness that influences these relational effects.

\subsection{Using clones of familiar voices configures expectations more than it transfers traits}

Interpreting the results of Studies 1 and 2 through the conceptual distinction introduced in section \ref{concepts} helps clarify the mechanisms at work. Participants frequently recognized or commented on the familiar voice, indicating that \textit{identity activation} occurred. The cloned voice triggered comparisons with the original speaker and shaped expectations about how the character should appear or behave.

Evidence for \textit{trait transfer} was considerably weaker. Only a small number of traits showed significant alignment effects, and these effects were inconsistent across contexts. Rather than systematically projecting personality characteristics from the original speaker to the artificial character, participants appeared to rely primarily on cues present in the stimulus itself.

Most importantly, the data provide little to no evidence for \textit{person reconstruction}. The familiar voice did not reinstate a broader person impression corresponding to the remembered individual. Instead, the cloned voice functioned as a recognizable cue that activated expectations while remaining embedded in a multimodal environment where other cues such as visual appearance and perceived artificiality strongly shaped evaluation.

Importantly, participants often recognized the speaker identity in the familiar voice condition. Their comments focused on appearance fit and comparison with the original person. This aligns with evidence that cloned familiar voices can pass categorical identity thresholds while still deviating from the richer stored prototype \cite{Fontaine.2017,Kanber.2022,Latinus.2011,Rosi.2025}. Recognition can be high even when synthetic origin is detected only modestly above chance. \textit{Identity activation} does not require reliable authenticity detection \cite{Barrington.2025}. Our data suggest that such activation mainly configures expectations about appearance and behavior rather than transferring the speaker's broader trait profile.

In self-voice paradigms, similarity-based mechanisms increase likability and behavioral trust, often via perceived homophily and personification \cite{Kuch.2025,Schanke.2022}. Third-party familiar voices introduce a stronger reference standard and invite prototype comparison. This amplifies scrutiny and highlights deviations in prosody, style, or interactional fit \cite{Fontaine.2017,Kanber.2022}. Under these conditions, a familiar voice may heighten mismatch salience without producing a reliable positive shift in applied evaluations.

\subsection{Boundary conditions suggested by the broader evidence}
In the introduction, we raised the ethical concern that cloned voices could be used to strategically shape how people evaluate artificial characters. Prior research suggests, however, that such effects depend strongly on the surrounding context. When audiovisual cues are inconsistent, trust and believability decrease and uncanny impressions increase \cite{Higgins.2022,Ruijten.2016}. In contrast, when voice, appearance, and interaction quality are coherent, similarity cues have more opportunity to shape evaluation because they are less likely to be overshadowed by other concerns \cite{Bodenhausen.2010,Kanber.2022,Rosi.2025}.

Our stimuli were intentionally controlled and relatively brief in order to isolate the effect of the voice cue. Even under these conditions, participants repeatedly reported that the available information was insufficient for reliable personality judgments. This suggests that short voice exposures alone may not provide enough information to support stable person-based projection. Rather than being inert, the effectiveness of known-voice cloning appears to depend on the broader cue environment. In particular, the visual appearance and behavioral coherence of the system deploying the voice may strongly constrain how the voice is interpreted. Alignment across multiple cues may therefore be required for familiar voices to meaningfully shape social evaluation.

\subsection{Ethical implications}
Ethically, these findings qualify rather than eliminate concerns about familiar voice cloning. The data do not support the assumption that a familiar cloned voice alone is sufficient to make an artificial character appear socially continuous with the original speaker. However, recognizable voices can still activate identity-based expectations and can make mismatches between voice, appearance, and behavior more salient.

This raises the question under what conditions cloned familiar voices should be used in social robotics and AI, if at all. What is it that a familiar voice could add to an artificial system that would enhance its capacities to function in a social context? One suggestion is that this use could increase the willingness to engage with systems meant to assist in educational and health practices. As we saw from our studies, however, the use of a familiar voice heightens awareness of the mismatch between the identity cue and the carrier of the voice. What is more, some participants stated that a mismatch between the voice and the carrier created uncanny or unsettling impressions (see section \ref{qualitativefindings}). Both of these findings make it likely that the use of a familiar voice would reduce the social functionality of the system rather than stimulate it. A further factor to consider is that especially in education and healthcare, users will be vulnerable. It might be that these users have more difficulty distinguishing reality from simulation, making it easier to manipulate them, or will be more affected by the uncanny or unsettling impressions. Given these factors, we therefore discourage the default use of familiar voices in social robotics, as well as their use without explicit consent. 

At the same time, we acknowledge that voice cloning can have beneficial uses as well, like providing autonomy to people with speech impediments or voice loss by allowing them to communicate in a voice that they identify with \cite{Mills.2014}. We can even imagine scenarios in which users have the option to give a particular voice to a robot that they personally own. In both of these cases, the autonomy of the user is protected by turning the use of a familiar cloned voice into a \textit{choice}. Even so, this should only be allowed under the condition that voice cloning itself is better regulated. To protect the autonomy of voice-owners, this entails developing transparent mechanisms to give consent for voice cloning, as well as the possibility to keep track of, update, and retract this consent. Additionally, the platforms that provide access to these voice cloning technologies should be properly monitored to ensure that they have these mechanisms in place. To protect the autonomy of the listeners, it is necessary to develop better mechanisms to detect impersonations of a person's voice, also known as \textit{deepfakes} \cite{blue2022}. Here, both non-technical mechanisms such as \textit{family/friend codewords} \cite{Sharevski.2025} and technical mechanisms such as deepfake detectors that differentiate between human and synthetic voices \cite{blue2022} are promising candidates. 

Finally, research into the effects of familiar voice cloning on personality transfer is bound to continue. We caution those pursuing this line of research to keep the following in mind:

\begin{itemize}
\item \textbf{Treat a cloned familiar voice as a weak identity cue.} Our results show no robust transfer of perceived personality, only small and selective alignment effects.
\item \textbf{Multimodal coherence matters.} Perceived artificiality and voice--appearance mismatch dominate judgments and override familiarity effects.
\item \textbf{Consider selective bias.} Even without broad personality transfer, recognizable voices can shift expectations and influence specific trait attributions.
\end{itemize}

\noindent Even when broad \textit{trait transfer} fails, recognizable voices can still activate expectations and subtly bias evaluations. The use of familiar voices in Human-Robot and AI interaction therefore requires transparency, consent, and careful attention to multimodal coherence.

\section{Limitations and Future Directions}
\label{limitations}

Several limitations should be considered. First, both studies relied on brief, single-exposure stimuli, approximately 14~s in Study~1 and 43~s in Study~2. This helped isolate the voice manipulation, but may have limited the formation of stable person-based impressions.

Second, participants sometimes perceived the cloned voices and virtual characters as artificial. Such cues may have drawn attention away from personality attribution and toward authenticity or audiovisual fit.

Third, generalizability is limited by the sample sizes and study contexts. Study~1 involved a small university sample evaluating a single lecturer, while Study~2 examined online communities around two content creators. The observed trait-selective effects therefore require replication.

Fourth, the studies focused on relatively specific forms of familiarity. Effects may differ for closer or more emotionally significant relationships, such as family members, close friends, or long-term colleagues. The findings are also limited to German-speaking contexts and explicit self-report measures.

Future work should examine familiar voice cloning in longer and more naturalistic interactions, across different forms of embodiment, and in voice-only settings. It should also compare different relationship types, include cross-linguistic and cross-cultural replications, and complement self-report ratings with behavioral or implicit measures. These extensions would help clarify when familiar voice cues meaningfully shape social evaluation and when their influence remains constrained by the broader interaction context.

\section{Conclusion}
\label{conclusion}

Voice cloning enables embodied characters to speak with the recognizable voice of a familiar person. Across two studies, we examined whether this familiarity leads listeners to transfer parts of their internal person model from the original speaker to the artificial character.

Our findings provide little evidence for robust personality transfer. Effects were selective and did not generalize across traits or study contexts. Viewed through the conceptual framework introduced in this paper, the implementation of a familiar voice in an embodied artificial character appears to produce \textit{identity activation} and, at most, limited \textit{trait transfer}, while providing little evidence for \textit{person reconstruction}.

These results suggest that familiar voice cloning functions less as a mechanism for reproducing personality and more as a context-dependent identity cue. Recognizable voices can shape expectations and selectively influence evaluations, yet their effects remain constrained by factors such as perceived artificiality and voice--appearance coherence.

From an ethical perspective, this moderates concerns that cloned familiar voices straightforwardly make artificial characters appear socially continuous with the individuals they resemble. At the same time, recognizable vocal identities can still activate trust-related expectations and subtly bias perception. Responsible deployment therefore requires clarity about the artificial system's status and should avoid implying personal continuity or endorsement by the original speaker.

\begin{credits}
\subsubsection{\ackname}
This research was funded by the European Union's Horizon Europe research and innovation programme under the CAVAA project (Grant Agreement No. 101071178), which deals with counterfactual assessment and valuation for an artificial awareness architecture, and the PILLAR-Robots project (Grant Agreement No. 101070381), which aims to develop purposeful intrinsically motivated lifelong learning autonomous robots.
\subsubsection{\discintname}
The authors have no competing interests to declare that are relevant to the content of this article.
\paragraph{Prior Version.}
A previous non-archival version of this work was presented as an extended abstract in the context of a non-archival poster \cite{Kuch.2026}. That version already reported both empirical studies, including methodological information. The present paper builds on that empirical foundation but revises the framing and emphasis. While the data and core findings remain the same, the contribution is recontextualized for the Robophilosophy audience by focusing more explicitly on the limits of identity cue transfer, the conceptual implications of voice-based person reference, and the ethical positioning of embodied artificial characters.
\end{credits}
%
%
%
\bibliographystyle{splncs04}
\raggedbottom
\bibliography{literature}

\clearpage

\end{document}